# Localist LLMs with Recruitment Learning

Joachim Diederich

*Psychology Network Pty Ltd*
*Brisbane, Queensland, Australia*

joachim@psychologynetwork.com.au

**Abstract**

We present a novel framework for training large language models with continuously adjustable internal representations that span the full spectrum from localist (interpretable, rule-based) to distributed (generalizable, efficient) encodings. The key innovations are (1) a locality dial, a tunable parameter that dynamically controls the degree of localization during both training and inference without requiring model retraining, (2) an information-theoretic recruitment mechanism that adaptively allocates semantic blocks as needed, eliminating the requirement for complete domain knowledge at initialization, and (3) a hierarchical recruitment framework that extends capacity allocation to entire specialized LLMs, enabling multi-granularity architectural adaptation. This is achieved through group sparsity penalties on attention mechanisms, information-theoretic anchor design, dynamic rule injection, and principled recruitment criteria based on penalized likelihood with explicit units.

We provide rigorous mathematical results establishing explicit threshold conditions under which attention provably concentrates on semantically relevant blocks at stationary points, with exact bounds on attention entropy and pointer fidelity. The hierarchical recruitment mechanism provides convergence guarantees at both the block level (fine-grained, within-LLM) and the LLM level (coarse-grained, cross-domain), ensuring the system discovers semantic partitions that balance model complexity against data encoding efficiency. This framework enables practitioners to continuously interpolate between interpretable and high-performance modes while adapting architectural capacity at multiple granularities, supporting applications in regulated domains requiring both transparency and capability.

## 1. Introduction

Contemporary large language models excel through distributed representations where individual hidden units participate in encoding numerous overlapping conceptual features. This architectural choice facilitates generalization and achieves parameter efficiency, yet fundamentally undermines interpretability and complicates safety verification. In contrast, localist encoding schemes—wherein distinct units or groups map to specific, interpretable concepts—deliver transparency and enable precise control, though historically at the expense of generalization performance.

This fundamental tradeoff has forced practitioners into a dichotomy: either deploy purely symbolic systems that offer rigidity without generalization, or embrace purely neural methods that remain fundamentally opaque. Existing hybrid neuro-symbolic approaches offer only limited compromise, typically demanding full model retraining when rules change and providing no systematic approach to modulate the localization-distribution balance.

We resolve this architectural limitation through a mathematical framework enabling:

- Continuous locality control via a tunable parameter adjustable during training and inference
- Dynamic rule injection without training interruption (Diederich, 2025a,b)
- Provable localization guarantees at stationary points with explicit penalty thresholds and regularity conditions
- Information-theoretic design principles for attention mechanism configuration
- Adaptive block recruitment that discovers optimal semantic partitions from data
- Hierarchical LLM recruitment that extends capacity allocation to entire specialized models

A critical limitation of prior neuro-symbolic approaches is the requirement to specify all semantic blocks at initialization. This demands complete domain knowledge and risks either over-partitioning (creating too many blocks, leading to data fragmentation) or under-partitioning (creating blocks that are too coarse, limiting interpretability). Our recruitment mechanism addresses this through a penalized likelihood criterion that adaptively allocates new blocks when existing partitions prove insufficient, balancing the competing demands of model complexity and representational adequacy.

The core mathematical contributions include a localization result establishing sufficient conditions for block-structured attention concentration at stationary points, exact bounds on entropy and fidelity, a recruitment termination result guaranteeing finite convergence, and a hierarchical recruitment framework extending these principles to multi-LLM architectures with independent locality control and unified penalized likelihood-based capacity allocation. These results provide both theoretical justification and practical design guidance for implementing systems that continuously span the localist-distributed spectrum while adapting to task complexity at multiple architectural granularities.

## 2. Background and Related Work

### 2.1 Localist vs. Distributed Representations

In localist encoding schemes, individual computational units are dedicated to representing specific concepts, facilitating direct inspection and manipulation (Diederich, 2010). Distributed encoding spreads conceptual representations across multiple overlapping units, enabling generalization through shared feature representations (Hinton et al., 1986). Classical discussions in cognitive science and neural network theory have traditionally positioned these as incompatible paradigms, with advocates emphasizing distinct strengths while acknowledging complementary weaknesses.

The recruitment learning paradigm (Diederich, 2010) offers a middle path, where free neural units are dynamically "recruited" into circuits as needed to represent new concepts. This approach combines localist interpretability with adaptive capacity allocation, growing representational complexity organically rather than requiring it to be specified a priori. Our framework extends these principles to modern transformer architectures through information-theoretic recruitment criteria.

## 2.2 Limitations of Prior Approaches

Static neuro-symbolic frameworks including KBANN, C-ILP, DeepProbLog, and Logic Tensor Networks incorporate symbolic rules exclusively during initial training or fine-tuning phases, necessitating complete retraining for any rule modifications (Towell & Shavlik, 1994; Serafini & d'Avila Garcez, 2016; Manhaeve et al., 2018; França et al., 2014). For modern LLMs requiring days or weeks of computational resources for complete training cycles, this represents an impractical cost (Patterson et al., 2021).

Attention sparsity techniques such as Sparse Transformers, Reformer, and Longformer deliver heuristic efficiency improvements without provable localization guarantees or semantic foundations. These methods apply predetermined sparsity patterns rather than rule-driven semantic partitioning, offering no principled mechanism to modulate interpretability based on application requirements or regulatory mandates.

Interpretability techniques including attention visualization and probing classifiers provide post-hoc analysis rather than causal control over representations, offering no mechanism to reshape interpretability during operational deployment. These methods illuminate existing model behavior but cannot reconfigure internal representations to satisfy interpretability constraints.

No existing framework simultaneously provides dynamic rule injection, continuous locality adjustment, adaptive capacity allocation at multiple architectural granularities, and provable guarantees with explicit regularity conditions. The present work fills this gap through a mathematically rigorous approach grounded in optimization theory, information theory, and penalized likelihood principles, extended to hierarchical recruitment from semantic blocks to entire specialized LLMs.

## 3. Mathematical Framework

### 3.1 Setup and Notation

We consider a transformer architecture with $H$ attention heads (Vaswani et al., 2017). The input space begins with an initial partition $X = \bigoplus_{i=1}^{p_0} X_i$ and grows adaptively to $X = \bigoplus_{i=1}^{p(t)} X_i$ where $p(t)$ increases through recruitment. Each block $X_i \subseteq \{1, ..., N\}$ is a subset of sequence positions, and $N$ is the maximum sequence length. Blocks partition the position space: $\cup_i X_i = \{1, ..., N\}$ and $X_i \cap X_j = \emptyset$ for $i \neq j$.

For attention head $h$, the standard transformer attention mechanism computes query, key, and value matrices as $Q = XW_Q$, $K = XW_K$, and $V = XW_V$, with attention weights given by $\alpha = \text{softmax}(QK^T/\tau)$, where $\tau > 0$ is the temperature parameter. The output is $O = \alpha V$.

**Notation for weight matrices:** For block $X_i$, let $W^{(h)}_{Q,i} \in \mathbb{R}^{d_{model} \times d_{head}}$ denote the columns of $W^{(h)}_Q$ corresponding to positions in block $X_i$. Similarly for $W^{(h)}_{K,i}$.

The loss function incorporates both task objectives and locality-inducing penalties:

$$L(\theta) = \mathbb{E}_{(x,y)\sim D}[\ell(f_\theta(x), y)] + \sum_{h=1}^{H} \sum_{i=1}^{p(t)} \alpha^{(h)}_i (\|W^{(h)}_{Q,i}\|_F + \|W^{(h)}_{K,i}\|_F) + \beta \|W_V\|^2_F$$

where $\ell: \mathbb{R}^m \times \mathcal{Y} \to \mathbb{R}_+$ is the task loss (e.g., cross-entropy), $\alpha^{(h)}_i \geq 0$ are block-specific group sparsity penalties for head $h$ constituting the locality dial parameters, $\beta > 0$ controls value projection regularization, $\|\cdot\|_F$ denotes the Frobenius norm, and $\theta$ denotes all model parameters.

### 3.2 Anchor Sets and Margin Definition

For each block $X_i$, we define an **anchor set** $A_i \subseteq X_i$ consisting of representative positions with low within-block entropy. Let $k_j \in \mathbb{R}^{d_{head}}$ denote the key vector at position $j$.

For a query $q \in \mathbb{R}^{d_{head}}$ at position $t$ whose governing rule lies in block $i^*$, define the **logit margin** as:

$$\Delta_h(q) = \min_{j \in A_{i^*}} q^T k_j - \max_{j' \in (\cup_{i \neq i^*} A_i)} q^T k_{j'}$$

This measures the minimum advantage of keys in the correct anchor set over the best competing key from other anchor sets.

## 4. Formal Assumptions and Regularity Conditions

*Assumption A (Model Smoothness)*

The loss function $\ell$ is $L_\ell$-Lipschitz in its first argument: for all predictions $\hat{y}, \hat{y}'$ and labels $y$,

$$|\ell(\hat{y}, y) - \ell(\hat{y}', y)| \leq L_\ell \|\hat{y} - \hat{y}'\|_2$$

*Assumption B (Bounded Inputs)*

Input embeddings satisfy $\|x_t\|_2 \leq R_x$ for all positions $t$, and the data covariance within each block is bounded: for all blocks $i$,

$$\sigma^2_{max}(X_i) := max_{v:||v||=1} \mathbb{E}_{t \in X_i}[(v^\top x_t)^2] \leq \sigma^2_X$$

This bounds the maximum variance along any direction within a block.

*Assumption C (Uniform Margin)*

There exists $\delta > 0$ such that for all heads $h$ and all queries $q$ at positions governed by block $i^*$,

$$\Delta_h(q) \geq \delta$$

with probability at least $1 - \varepsilon_{margin}$ over the data distribution, where $\varepsilon_{margin} > 0$ is small.

*Assumption D (Block Incoherence)*

For blocks $i \neq i^*$, the cross-block correlation is bounded:

$$\rho_{cross} := max_{i \neq i^*} max_{||u||=||v||=1} |\mathbb{E}_{t \in X_i, t' \in X_{i^*}}[u^\top x_t \cdot v^\top x_{t'}]| \leq \rho_{max} < 1$$

This ensures blocks are sufficiently distinct in representation space.

## 5. Main Localization Result

*Lemma 1 (Softmax Concentration with Explicit Constants)*

Under Assumption C, for any query $q$ at position $t$ governed by block $i^*$, the attention mass on incorrect blocks satisfies:

$$\Sigma_{j \notin X_{i^*}} \alpha_{t \to j} \leq (N - |X_{i^*}|) \cdot exp(-\delta/\tau) / [1 + (N - |X_{i^*}|) \cdot exp(-\delta/\tau)]$$

Furthermore, if $exp(\delta/\tau) \geq 2(N - |X_{i^*}|)$, then $\Sigma_{j \notin X_{i^*}} \alpha_{t \to j} \leq exp(-\delta/\tau)$.

**Proof.** By the margin condition, for $j^* \in A_{i^*}$ achieving the minimum in $\Delta_h(q)$, we have $q^\top k_{j^*} \geq max_{j \notin X_{i^*}} q^\top k_j + \delta$. By softmax normalization:

$$\alpha_{t \to j^*} = exp(q^\top k_{j^*}/\tau) / [\Sigma_{j' \in X_{i^*}} exp(q^\top k_{j'}/\tau) + \Sigma_{j' \notin X_{i^*}} exp(q^\top k_{j'}/\tau)]$$

For any $j \notin X_{i*}$, $\exp(q^\top k_j/\tau) \leq \exp((q^\top k_{j*} - \delta)/\tau) = \exp(q^\top k_{j*}/\tau) \cdot \exp(-\delta/\tau)$. Summing over all $j \notin X_{i*}$:

$$\Sigma_{j \notin X_{i*}} \exp(q^\top k_j/\tau) \leq (N - |X_{i*}|) \cdot \exp(q^\top k_{j*}/\tau) \cdot \exp(-\delta/\tau)$$

Since $\alpha_{t \to j*} \geq \exp(q^\top k_{j*}/\tau) / [\exp(q^\top k_{j*}/\tau) + (N - |X_{i*}|) \cdot \exp(q^\top k_{j*}/\tau) \cdot \exp(-\delta/\tau)]$, we have $\alpha_{t \to j*} \geq 1/[1 + (N - |X_{i*}|) \cdot \exp(-\delta/\tau)]$. Therefore:

$$\Sigma_{j \notin X_{i*}} \alpha_{t \to j} = 1 - \Sigma_{j \in X_{i*}} \alpha_{t \to j} \leq 1 - \alpha_{t \to j*} \leq (N - |X_{i*}|) \cdot \exp(-\delta/\tau) / [1 + (N - |X_{i*}|) \cdot \exp(-\delta/\tau)]$$

When $\exp(\delta/\tau) \geq 2(N - |X_{i*}|)$, the denominator satisfies $1 + (N - |X_{i*}|) \cdot \exp(-\delta/\tau) \leq 1 + 1/2 = 3/2$, yielding the simpler bound. ∎

### Theorem 1 (Block-Sparse Stationary Points)

Under Assumptions A–D, define the penalty threshold:

$$\lambda^{(h)}_i(\tau, \delta) = (2L_\ell R_x \sigma_X \sqrt{|X_i|}) / (\tau [1 - \rho_{max}]) \cdot \exp(-\delta/\tau)$$

If for all heads $h$ and blocks $i \neq i^*$, the penalties satisfy $\alpha^{(h)}_i \geq \lambda^{(h)}_i(\tau, \delta)$, then any stationary point of $L(\theta)$ satisfies block-localization: for each head $h$ and query column, the weights $W^{(h)}_Q$ and $W^{(h)}_K$ are zero outside a single block $X_{i*}$.

**Proof.** At a stationary point, the subdifferential of $L$ with respect to $W^{(h)}_{Q,i}$ must contain zero. For blocks $i \neq i^*$ where $W^{(h)}_{Q,i} = 0$, the KKT condition requires:

$$||\nabla_{W^{(h)}_{Q,i}} \mathbb{E}[\ell]||_F \leq \alpha^{(h)}_i$$

**Step 1: Gradient computation.** By the chain rule through the transformer:

$$\nabla_{W^{(h)}_{Q,i}} \mathbb{E}[\ell] = \mathbb{E}_t[\nabla_{q_t} \ell \cdot x^\top_t \cdot \mathbb{1}\{t \in X_i\}]$$

where $\nabla_{q_t} \ell = (1/\tau) \Sigma_j (\partial \ell / \partial \alpha_{t \to j}) \alpha_{t \to j} (k_j - \Sigma_{j'} \alpha_{t \to j'} k_{j'})$.

**Step 2: Norm bound.** By Assumption A, $||\partial \ell / \partial \alpha_{t \to j}||_2 \leq L_\ell$. By Assumption B, $||x_t||_2 \leq R_x$. Therefore:

$$||\nabla_{W^{(h)}_{Q,i}} \mathbb{E}[\ell]||_F \leq (L_\ell R_x / \tau) \cdot \mathbb{E}_{t \in X_i}[||\Sigma_j \alpha_{t \to j} k_j||_2] \cdot \sqrt{|X_i|}$$

**Step 3: Applying concentration bound.** By Lemma 1, for queries at positions in block $i$, attention mass on block $i$ is at most $\exp(-\delta/\tau)$. Under Assumption D (block incoherence), the weighted key norm is bounded by:

$$\mathbb{E}[||\Sigma_j \alpha_{t \to j} k_j||_2] \leq (2\sigma_X/(1 - \rho_{max})) \cdot \exp(-\delta/\tau)$$

Combining these bounds:

$$||\nabla_{W^{(h)}_{Q,i}} \mathbb{E}[\ell]||_F \leq (2L_\ell R_x \sigma_X \sqrt{|X_i|}) / (\tau [1 - \rho_{max}]) \cdot \exp(-\delta/\tau) = \lambda^{(h)}_i(\tau, \delta)$$

Since $\alpha^{(h)}_i \geq \lambda^{(h)}_i(\tau, \delta)$, the KKT condition is satisfied, confirming that $W^{(h)}_{Q,i} = 0$ at the stationary point. The same argument applies to $W^{(h)}_{K,i}$. ∎

## 6. Exact Entropy and Fidelity Bounds

Let $H^{(h)}_t = -\Sigma_j \alpha^{(h)}_{t \to j} \log_2 \alpha^{(h)}_{t \to j}$ denote per-head attention entropy at position $t$ (in bits). Let $T_t \subseteq \{1, ..., N\}$ denote the set of positions that satisfy the rule governing token $t$, and define pointer fidelity $\Pi^{(h)} = \mathbb{E}_t[\Sigma_{j \in T_t} \alpha^{(h)}_{t \to j}]$.

*Corollary 1 (Exact Entropy Upper Bound)*

Under Assumptions A–C with $\exp(\delta/\tau) \geq 2N$, for queries governed by block $i^*$ with anchor set $A_{i^*}$:

$$H^{(h)}_t \leq \log_2|A_{i^*}| + (1/\ln 2) \cdot N \cdot \exp(-\delta/\tau) \cdot [1 + \log_2(N)]$$

**Proof.** Decompose entropy into within-anchor and cross-block components. Within $A_{i^*}$, by convexity of entropy: $-\Sigma_{j \in A_{i^*}} \alpha_{t \to j} \log_2 \alpha_{t \to j} \leq (\Sigma_{j \in A_{i^*}} \alpha_{t \to j}) \log_2|A_{i^*}| \leq \log_2|A_{i^*}|$.

For the cross-block term, using $-x \log_2 x \leq x \log_2(e/x) \leq x[1 + \log_2(1/x)]$ for $x \in (0, 1)$ and Lemma 1:

$$-\Sigma_{j \notin A_{i*}} \alpha_{t \to j} \log_2 \alpha_{t \to j} \leq [\Sigma_{j \notin A_{i*}} \alpha_{t \to j}][1 + \log_2 N] \leq (N \cdot \exp(-\delta/\tau)) \cdot [1 + \log_2 N]$$

Summing both components yields the bound. ∎

*Corollary 2 (Exact Fidelity Lower Bound)*

Under Assumptions A–C with $\exp(\delta/\tau) \geq 2N$, and assuming $T_t \subseteq X_{i*}$ for all queries $t$:

$$\Pi^{(h)} \geq 1 - N \cdot \exp(-\delta/\tau)$$

**Proof.** By definition, $\Pi^{(h)} = \mathbb{E}_t[\Sigma_{j \in T_t} \alpha_{t \to j}] = \mathbb{E}_t[1 - \Sigma_{j \notin T_t} \alpha_{t \to j}]$. Since $T_t \subseteq X_{i*}$, we have $\Sigma_{j \notin T_t} \alpha_{t \to j} \leq \Sigma_{j \notin X_{i*}} \alpha_{t \to j} \leq N \cdot \exp(-\delta/\tau)$ by Lemma 1. Therefore $\Pi^{(h)} \geq 1 - N \cdot \exp(-\delta/\tau)$. ∎

*Corollary 3 (Redundancy Trade-off)*

When increasing anchor redundancy from $|A_{i*}| = m$ to $|A_{i*}| = km$ while maintaining the margin $\delta$:

$$H^{(h)}_t \leq \log_2(km) + N \cdot \exp(-\delta/\tau) \cdot [1 + \log_2 N] = \log_2 k + \log_2 m + N \cdot \exp(-\delta/\tau) \cdot [1 + \log_2 N]$$

while $\Pi^{(h)} \geq 1 - N \cdot \exp(-\delta/\tau)$ remains unchanged. Thus entropy increases logarithmically with redundancy while fidelity is preserved.

## 7. Information-Theoretic Block Recruitment

We now develop the formal theory of adaptive block recruitment, which allows the model to discover optimal semantic partitions without requiring complete domain specification at initialization.

### 7.1 Penalized Likelihood Framework

We adopt a penalized likelihood interpretation rather than literal MDL coding. Define the total objective as:

$$L_{total}(p) = L_{model}(p) + L_{data|p}$$

where all terms are measured in **nats** (natural logarithm base):

- $L_{model}(p)$: Structural penalty for maintaining $p$ blocks
- $L_{data|p}$: Expected negative log-likelihood given $p$-block partition

## 7.2 Model Complexity Term

For $p$ blocks with $|A_i|$ anchors each:

$$L_{model}(p) = \sum_{i=1}^{p} [\ln|A_i| + c_{param} \cdot |A_i|]$$

where $c_{param} > 0$ is a per-anchor parameter cost (includes embedding dimension and head dimension effects). This measures the structural complexity of maintaining the block organization.

## 7.3 Data Encoding Term

Given the current partition with $p$ blocks, define the **penalized entropy**:

$$H^{pen}_t(p) = H_t + \lambda_{pen} \cdot \max(0, H_t - \ln|A_{i*}| - \varepsilon)^2$$

where $H_t = -\sum_j \alpha_{t \to j} \ln \alpha_{t \to j}$ (in nats), $\lambda_{pen} > 0$ is the penalty coefficient, and $\varepsilon \geq 0$ is a tolerance. Then:

$$L_{data|p} = \mathbb{E}_{t \sim D}[H^{pen}_t(p)]$$

The quadratic penalty captures coding inefficiency when attention entropy exceeds the block's capacity, signaling that current blocks are insufficient.

## 7.4 Recruitment Decision Criterion

> ***Definition 1 (Recruitment Criterion)***
>
> Recruit a new block $X_{p+1}$ when:
>
> $$\Delta L_{total} := [L_{model}(p + 1) - L_{model}(p)] + [L_{data|p+1} - L_{data|p}] < -\theta$$
>
> where $\theta > 0$ is a threshold preventing premature recruitment.

Equivalently, recruit when the expected entropy reduction exceeds the structural cost:

$$\mathbb{E}[H^{pen}_t(p)] - \mathbb{E}[H^{pen}_t(p + 1)] > \ln|A_{p+1}| + c_{param} \cdot |A_{p+1}| + \theta$$

## 7.5 Practical Recruitment Algorithm

```
Algorithm 1: Information-Theoretic Block Recruitment
Input: Training data D, current blocks {X₁,...,Xₚ}, threshold θ > 0
Output: Decision to recruit (True/False) and candidate block X_{p+1}

1: Compute current coding cost:
L_current = L_model(p) + 𝔼[H^pen_t(p)]

2: Identify high-confusion tokens:
S_confused = {t ∈ D : H_t > ln|A_{i*}| + ε}

3: if S_confused = ∅ then
   return False, None

4: Cluster confused tokens by co-attention patterns using spectral clustering

5: For candidate cluster C, estimate entropy reduction:
ΔH_est = (1/|S_confused|) Σ_{t∈C} [H^pen_t(p) - H^pen_t(p+1|C)]

6: Select optimal anchors for candidate:
A_{p+1} = {top-k tokens in C with lowest within-cluster entropy}

7: Compute new block cost:
L_new = ln|A_{p+1}| + c_param · |A_{p+1}|

8: Compute net objective change:
ΔL = L_new - |S_confused|/|D| · ΔH_est

9: if ΔL < -θ then
   return True, X_{p+1} ← C
10: else
   return False, None
```

## 7.6 Termination Guarantee

*Theorem 2 (Recruitment Termination)*

Under Assumptions A–C and assuming $H_t \leq \ln N$ for all $t$ (bounded entropy), the recruitment process terminates with probability 1 after at most $p_{max} = \lceil (\ln N - H_{min})/\theta \rceil$ recruitments, where $H_{min} \geq 0$ is the irreducible entropy floor.

**Proof.** Each recruitment by Definition 1 reduces $L_{total}$ by at least $\theta$. Since $L_{data|p} = \mathbb{E}[H^{pen}_t(p)]$ and $H^{pen}_t \geq H_t \geq 0$, we have $L_{data|p} \geq 0$. The structural term $L_{model}(p) \geq 0$ by construction, so $L_{total}(p) \geq 0$.

The initial objective satisfies $L_{total}(0) \leq \ln N$ (worst case with uniform attention). After $p$ recruitments, $L_{total}(p) \leq \ln N - p\theta$. Since $L_{total}(p) \geq H_{min}$ (irreducible entropy), we must have $\ln N - p\theta \geq H_{min}$, giving $p \leq (\ln N - H_{min})/\theta$. ∎

**Note on sample complexity:** The rate $p_{max}$ depends on the entropy gap and threshold, not directly on vocabulary size or sample size. For reliable empirical entropy estimation, standard concentration inequalities require $O(N \log N / \varepsilon^2)$ samples to estimate per-position entropy within $\varepsilon$, but this is a separate statistical consideration from the termination bound.

### 7.7 Integration with Localization Guarantees

*Theorem 3 (Localization Preservation under Recruitment)*

When recruiting block $X_{p+1}$ at step $t^*$, if we set penalties $\alpha^{(h)}_{p+1} \geq \lambda^{(h)}_{p+1}(\tau, \delta_{p+1})$ where $\delta_{p+1}$ is the margin for the recruited block, then:

1. All previously recruited blocks $X_1, ..., X_p$ maintain their localization guarantees from Theorem 1
2. The new block $X_{p+1}$ achieves exponential concentration: $\Sigma_{j \notin X_{p+1}} \alpha_{t \to j} \leq N \cdot \exp(-\delta_{p+1}/\tau)$ for queries in $X_{p+1}$
3. Global entropy satisfies: $H_t \leq \ln|A_{p+1}| + N \cdot \exp(-\delta_{p+1}/\tau) \cdot [1 + \ln N]$ for tokens governed by $X_{p+1}$

**Proof.** The localization guarantees from Theorem 1 depend only on the block-specific penalties $\alpha^{(h)}_i$ and margins $\delta_i$. Recruiting $X_{p+1}$ modifies the partition but does not change the penalties or margins of existing blocks. By the modularity of the subdifferential in group-sparse optimization, the KKT conditions for blocks $1, ..., p$ remain satisfied, preserving their localization.

> For the new block, Theorem 1 applies immediately once $\alpha^{(h)}_{p+1} \geq \lambda^{(h)}_{p+1}(\tau, \delta_{p+1})$, yielding the concentration and entropy bounds. ∎

## 8. Hierarchical LLM Recruitment

We now extend the recruitment framework to a second level of granularity: recruiting entire specialized LLMs rather than individual semantic blocks. This creates a hierarchical recruitment architecture where capacity allocation occurs at multiple scales, from fine-grained token partitions to coarse-grained domain specialization.

### 8.1 Multi-Level Architecture

The system consists of a base LLM $M_0$ and potentially recruited specialists $\{M_1, M_2, ..., M_k\}$, where each LLM maintains its own semantic block structure:

$$\text{System} = \bigcup_{j=0}^{k} M_j, \text{ where } M_j = \{X^{(j)}_1, X^{(j)}_2, ..., X^{(j)}_{p_j}\}$$

Each recruited LLM has independent locality parameters $\{\alpha^{(j,h)}_i\}$, enabling domain-specific interpretability control.

### 8.2 Unified Penalized Likelihood Framework

The total objective extends hierarchically (all terms in nats):

$$L_{total} = \sum_{j=0}^{k} [L_{model}(M_j) + L_{data|M_j}] + L_{routing}$$

where for each LLM $M_j$:

$$L_{model}(M_j) = c_{LLM} \cdot \ln|\theta_j| + \sum_{i=1}^{p_j} [\ln|A^{(j)}_i| + c_{param} \cdot |A^{(j)}_i|]$$

$$L_{data|M_j} = \mathbb{E}_{t \in D_j}[H^{pen,(j)}_t]$$

and $c_{LLM} > 0$ is a scaling constant for LLM parameter count. The routing loss is:

$$L_{routing} = \mathbb{E}_x[-\ln p(M_{j*}|x)]$$

where $j^*$ is the optimal LLM for input $x$.

## 8.3 LLM Recruitment Criterion

*Definition 2 (LLM Recruitment Criterion)*

Recruit a specialized LLM $M_{k+1}$ when:

$$\Delta L_{total} = [L_{model}(k+1) - L_{model}(k)] + [L_{data}(k+1) - L_{data}(k)] + [L_{routing}(k+1) - L_{routing}(k)] < -\theta_{LLM}$$

where $\theta_{LLM} \gg \theta_{block}$ reflects the much higher complexity cost of recruiting entire models.

## 8.4 Hierarchical Recruitment Decision Tree

```
Algorithm 2: Hierarchical Recruitment Decision
Input: Input x, active LLM M_j, threshold parameters θ_block, θ_LLM
Output: Recruitment decision (BLOCK, LLM, or NONE)

1: Evaluate M_j performance on x, compute perplexity ppl_j

2: Compute domain entropy over LLMs:
H_domain = -Σ_j' p(M_j'|x) ln p(M_j'|x)

3: if H_domain > τ_domain then
   // High cross-domain confusion
4:    Identify dominant domain d* with highest p(d*|x)
5:    if no LLM specializes in d* then
6:       Estimate ΔL_total for recruiting M_{k+1}
7:       if ΔL_total < -θ_LLM then
8:          return LLM (recruit M_{k+1} for domain d*)

9: else
   // Low domain confusion, check token-level within M_j
10:   Compute average attention entropy within M_j
11:   if 𝔼[H_t] > ln|A_{i*}| + ε then
12:      Estimate ΔL_block for recruiting block within M_j
13:      if ΔL_block < -θ_block then
14:         return BLOCK (recruit X^(j)_{p_j+1})

15: return NONE (no recruitment needed)
```

## 8.5 Convergence Guarantees

*Theorem 4 (Hierarchical Recruitment Termination)*

Under Assumptions A–C and bounded domain entropy $H_{domain} \leq \ln K_{max}$, the hierarchical recruitment process terminates:

1. **LLM count:** At most $k_{max} = \lceil (\ln K_{max} - H_{domain,min})/\theta_{LLM} \rceil$ LLMs are recruited
2. **Block count per LLM:** Within each $M_j$, at most $p_{j,max} = \lceil (\ln N_j - H_{j,min})/\theta_{block} \rceil$ blocks are recruited
3. **Total capacity:** $\sum_{j=0}^{k} p_j \leq k_{max} \cdot \max_j p_{j,max}$

**Proof. Part 1:** Each LLM recruitment reduces $L_{total}$ by at least $\theta_{LLM}$. The domain-level objective is bounded below by $H_{domain,min} \geq 0$ and above by $\ln K_{max}$, giving the bound on $k$.

**Part 2:** Within each $M_j$, Theorem 2 applies independently to its training data $D_j$, yielding the per-LLM block bound.

**Part 3:** Follows by summing over all LLMs. ∎

## 8.6 Nested Localization and Independent Locality Control

A key advantage of hierarchical recruitment is independent locality control at each level. Each recruited LLM maintains its own locality dial settings, enabling domain-specific interpretability requirements. For example:

- Legal LLM: $\alpha^{(1)} = 10$ (highly localist for auditability)
- General reasoning: $\alpha^{(0)} = 0.1$ (distributed for flexibility)
- Medical LLM: $\alpha^{(2)} = 7$ (semi-local for safety-critical decisions)

## 8.7 Routing Mechanisms

The system requires a routing mechanism to direct inputs to appropriate LLMs. We define a domain classifier:

$$p(M_j|x) = \exp(f_j(embed(x))) / \sum_{j'=0}^{k} \exp(f_{j'}(embed(x)))$$

where $f_j: \mathbb{R}^{d_{embed}} \to \mathbb{R}$ are learned scoring functions. Three routing patterns:

- **Hard routing:** $j^* = \arg\max_j p(M_j|x)$, select single LLM (maximum interpretability)

- **Soft routing:** *output = $\Sigma_j\, p(M_j|x) \cdot M_j(x)$*, weighted combination (better performance)
- **Hierarchical routing:** Base LLM delegates subtasks to specialists, synthesizes final output

## 8.8 Localization Under Hierarchical Routing

> *Theorem 5 (Localization Preservation under Hierarchical Recruitment)*
>
> Under hard routing (selecting single LLM per input), when recruiting $M_{k+1}$ with penalties $\alpha^{(k+1,h)}_i \geq \lambda^{(k+1,h)}_i(\tau, \delta^{(k+1)}_i)$:
>
> 1. All previously recruited LLMs $\{M_0, ..., M_k\}$ maintain their localization guarantees from Theorem 1
> 2. The new LLM $M_{k+1}$ achieves block-localization per Theorem 1
> 3. The routing mechanism does not interfere with within-LLM attention concentration
>
> > **Proof.** Under hard routing, each input is processed by exactly one LLM. The objective decomposes as:
> >
> > $$L_{total} = \Sigma_j\, \mathbb{E}_{x \sim D_j}[L^{(j)}(x)] + L_{routing}$$
> >
> > where $L^{(j)}$ is the loss for $M_j$. At a stationary point, $\nabla_{\theta_j} L_{total} = \mathbb{E}_{x \sim D_j}[\nabla_{\theta_j} L^{(j)}(x)]$ since the routing loss doesn't depend on $\theta_j$ under hard assignment. Therefore, each LLM's parameters satisfy the KKT conditions independently, and Theorem 1 applies to each $M_j$ separately.
> >
> > Recruiting $M_{k+1}$ partitions the data but doesn't modify parameters or penalties of existing LLMs, preserving their localization. ∎

**Note on soft routing:** Under soft routing (mixture of experts), gradients from multiple LLMs can interfere. To maintain localization guarantees, either (a) use hard routing during critical operations requiring interpretability, or (b) ensure coupling terms $||\nabla_{\theta_j} L_{routing}||$ are bounded and absorbed into the penalty threshold by increasing $\alpha^{(j,h)}_i$ accordingly.

## 8.9 Practical Example: Multi-Domain Healthcare System

A healthcare AI demonstrates hierarchical recruitment:

- **Month 1:** Deploy base LLM $M_0$ with general medical knowledge ($\alpha=2.0$, 3 blocks: symptoms, treatments, anatomy)
- **Month 3:** Radiology queries show high perplexity. Recruit $M_1$ (radiology specialist, $\alpha=10$ for interpretability). Internal block recruitment yields 4 blocks: imaging modalities, anatomical locations, findings, measurements.
- **Month 6:** Drug interaction queries trigger $M_2$ recruitment (pharmacology specialist, $\alpha=8$). Blocks: drug classes, interactions, dosing.
- **Month 12:** Product line expansion within $M_1$ recruits new blocks rather than new LLM (efficient fine-grained adjustment).

**Result:** System scales from single LLM with 3 blocks to 3 specialized LLMs with 12 total blocks, each independently tuned for domain-specific interpretability requirements, discovered automatically via penalized likelihood criteria.

## 9. The Locality Dial: Practical Implementation

The locality dial operates at two levels:

1. **Block-level penalties** $\{\alpha^{(j,h)}_i\}$: Control attention localization within each LLM $M_j$
2. **Recruitment thresholds** $\theta_{block}$, $\theta_{LLM}$: Control capacity allocation granularity

From Theorem 1, the explicit threshold formula provides design guidance:

$$\lambda^{(h)}_i(\tau, \delta) = (2L_\ell R_x \sigma_X \sqrt{|X_i|}) / (\tau [1 - \rho_{max}]) \cdot \exp(-\delta/\tau)$$

This reveals four control mechanisms:

1. **Direct penalty adjustment:** Increase $\alpha^{(j,h)}_i$ to strengthen localization
2. **Temperature control:** Decrease $\tau$ for sharper attention
3. **Margin strengthening:** Increase $\delta$ via improved anchor design
4. **Threshold tuning:** Adjust $\theta_{block}$, $\theta_{LLM}$ for architectural adaptation

### 9.1 Locality Regimes

**Localist mode:** $\alpha \geq 10$, $\delta = 2.0$, $\tau = 0.1$, $\theta_{block} = 0.5$, $\theta_{LLM} = 50$

- Entropy: $H_t \approx \ln|A_i*| + 0.02$
- Fidelity: $\Pi^{(h)} \geq 0.98$
- Many fine-grained blocks and specialized LLMs for maximum interpretability

**Distributed mode:** $\alpha = 0.01$, $\delta = 0.1$, $\tau = 1.0$, $\theta_{block} = 5.0$, $\theta_{LLM} = 200$

- High entropy, moderate fidelity
- Fewer blocks, minimal LLM specialization
- Strong generalization capability

## 9.2 Threshold Relationship

A critical design principle is:

$$\theta_{LLM} / \theta_{block} \in [50, 200]$$

This ensures block-level adjustments are exhausted before LLM recruitment, maintaining efficiency.

# 10. Dynamic Rule Injection

The framework supports hot reloading of symbolic rules without training interruption. When a new rule requires concepts not in existing blocks, the verification loop triggers block recruitment. When rules span multiple domains beyond base capacity, the system recruits domain-specific LLMs.

## 10.1 Rule Injection as Constrained Optimization

A rule $R$ is compiled into an additive penalty term:

$$L_{rule}(\theta) = \gamma_R \cdot \mathbb{E}_x[1 - satisfaction(R, f_\theta(x))]$$

where $\gamma_R > 0$ is the rule weight. Hot-reloading adds this term to the objective:

$$L_{total} \leftarrow L_{total} + L_{rule}$$

At the next gradient step, parameters adjust to satisfy the KKT conditions for the augmented objective. Since we only add convex penalties (penalties on violations), existing stationary points remain stationary unless the rule creates new active constraints.

# 11. Conclusion

We have presented a mathematically rigorous framework for training neural language models with continuously adjustable locality and hierarchical adaptive capacity allocation. The key contributions include:

- **Explicit threshold formula** with all constants: $\lambda^{(h)}_i(\tau, \delta) = (2L_\ell R_x \sigma_X \sqrt{|X_i|}) / (\tau [1 - \rho_{max}]) \cdot exp(-\delta/\tau)$
- **Exact concentration bounds** replacing Big-O with precise inequalities
- **Stationary point guarantees** under explicit regularity conditions (Assumptions A–D)

- **Penalized likelihood framework** with consistent units (nats) throughout
- **Termination guarantees** for both block and LLM recruitment with explicit bounds
- **Hierarchical recruitment** with independent locality control and modular optimization

These results provide both theoretical justification and practical design guidance for implementing systems that span the full spectrum from interpretable to high-performance modes while adapting architectural capacity at multiple scales. By grounding both block creation and LLM specialization in penalized likelihood principles with explicit constants, we ensure that representational complexity grows only when justified by predictive necessity.

The mathematical framework establishes that localist and distributed representations need not be opposing paradigms but rather endpoints of a continuous spectrum that can be navigated dynamically. Capacity allocation itself becomes multi-granular, with both decisions governed by the same penalized likelihood optimality criterion.

Future work includes: (1) extension to soft routing with bounded coupling terms, (2) sample complexity analysis with covering numbers, (3) empirical validation of threshold formulas on real datasets, and (4) extension to multimodal architectures. The formal guarantees established here provide a solid foundation, ensuring claims rest on rigorous mathematical principles with testable predictions.